\pdfoutput=1

\documentclass[11pt]{article}

\usepackage{acl}

\usepackage{times}
\usepackage{latexsym}

\usepackage[T1]{fontenc}

\usepackage[utf8]{inputenc}

\usepackage{microtype}

\usepackage{amsmath} 
\usepackage{amssymb}
\usepackage{graphicx}
\usepackage{booktabs} 
\usepackage{multirow} 
\usepackage{microtype}
\usepackage{graphicx}
\usepackage{multirow}
\usepackage{hyperref}
\usepackage{amsfonts}
\usepackage{bbding}
\usepackage{color}

\title{UnifiedMLLM: Enabling Unified Representation for Multi-modal Multi-tasks With Large Language Model}

\author{\textbf{Zhaowei Li\textsuperscript{\rm 1,2}},
    ~\textbf{Wei Wang\textsuperscript{\rm 1,3}}, 
    ~\textbf{Yiqing Cai\textsuperscript{\rm 1}},
    ~\textbf{Qi Xu\textsuperscript{\rm 1}},
    ~\textbf{Pengyu Wang\textsuperscript{\rm 2}},\\
    ~\textbf{Dong Zhang\textsuperscript{\rm 2}},
    ~\textbf{Hang Song\textsuperscript{\rm 1}},
    ~\textbf{Botian Jiang\textsuperscript{\rm 2}},
    ~\textbf{Zhida Huang\textsuperscript{\rm 1}},
    ~\textbf{Tao Wang\textsuperscript{\rm 1}}\\
    \textsuperscript{\rm 1}ByteDance Inc, \textsuperscript{\rm 2}Fudan University,\\ \textsuperscript{\rm 3}University of Science and Technology of China\\
    {\tt lizhaowei126@gmail.com} \\
}

\begin{document}
\maketitle
\begin{abstract}
Significant advancements has recently been achieved in the field of multi-modal large language models (MLLMs), demonstrating their remarkable capabilities in understanding and reasoning across diverse tasks. However, these models are often trained for specific tasks and rely on task-specific input-output formats, limiting their applicability to a broader range of tasks. 
This raises a fundamental question: Can we develop a unified approach to represent and handle different multi-modal tasks to maximize the generalizability of MLLMs? 
In this paper, we propose UnifiedMLLM, a comprehensive model designed to represent various tasks using a unified representation. 
Our model exhibits strong capabilities in comprehending the implicit intent of user instructions and preforming reasoning. In addition to generating textual responses, our model also outputs task tokens and grounding tokens, serving as indicators of task types and task granularity. These outputs are subsequently routed through the task router and directed to specific expert models for task completion.
To train our model, we construct a task-specific dataset and an 100k multi-task dataset encompassing complex scenarios. Employing a three-stage training strategy, we equip our model with robust reasoning and task processing capabilities while preserving its generalization capacity and knowledge reservoir.
Extensive experiments showcase the impressive performance of our unified representation approach across various tasks, surpassing existing methodologies.
Furthermore, our approach exhibits exceptional scalability and generality.
Our code, model, and dataset will be available at \url{https://github.com/lzw-lzw/UnifiedMLLM}.
\end{abstract}

\begin{figure*}[t]
    \centering
    \includegraphics[width=0.95\textwidth]{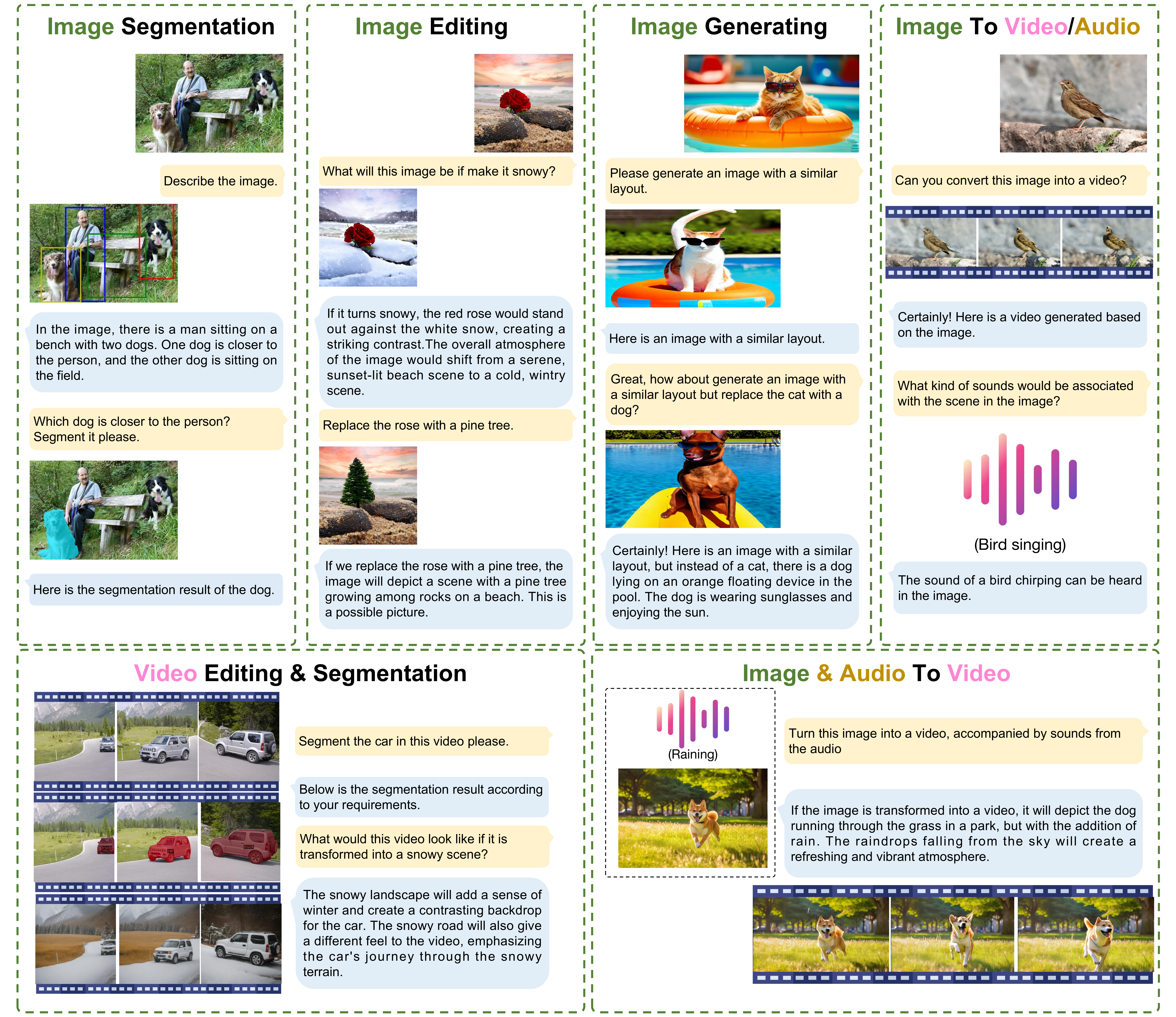}
    \caption{We introduce UnifiedMLLM, a model that represents and handles different multi-modal tasks in a unified manner, enabling it to perform tasks involving multi-modal understanding, processing, and generation.}
    \label{fig:demo}
\end{figure*}

\section{Introduction}
Large language models have demonstrated remarkable performance in various natural language processing tasks, and the field of multi-modal large language models (MLLMs) has also made significant progress. 
Representative models like LLaVA~\cite{liu2023visual} and MiniGPT-4~\cite{zhu2023minigpt} have exhibited great capabilities in tasks such as image captioning and visual question answering.
Some models have been designed to tackle a broader range of multi-modal tasks, including image segmentation~\cite{lai2023lisa} and image editing~\cite{huang2023smartedit} using MLLMs. However, these models are primarily designed and trained for specific tasks, which constrains their applicability to a broader range of tasks and their overall generality in diverse scenarios due to their reliance on task-specific input-output formats.
Some approaches have explored the utilization of MLLMs to accomplish more tasks. For example, LLaVA-Interactive~\cite{chen2023llava} integrates multiple visual expert models with LLaVA to perform tasks such as image segmentation, editing, generation. 

However, these methods view MLLMs as chatbots and heavily rely on scheduling expert models to handling visual tasks, thus failing to fully leverage the knowledge base and reasoning capabilities of MLLMs. 
Furthermore, while these models can handle multiple visual tasks simultaneously, they often rely on explicit instructions or predefined categories to execute visual tasks, lacking the ability to understand more implicit and complex human instructions. 
We expect models to comprehend implicit human intent, which encompasses understanding the tasks intend to perform and the specific regions where these tasks need to be executed. Therefore, it is necessary for models to possess strong reasoning and grounding abilities, which were lacking in previous work.

In this paper, we propose UnifiedMLLM, which models and handles different multi-modal tasks in a unified manner. 
Our approach introduce task tokens and grounding tokens to establish a unified representation across different tasks. 
The model understands the implicit intent behind user instructions and outputs not only the textual response but also our expanded special tokens indicating the task type and specific region to be processed. These tokens are then routed through a task router, activating the corresponding expert model for task execution.
Based on this design, as illustrated in Figure~\ref{fig:demo}, our model exhibits excellent performance in accomplishing a wide range of multi-modal tasks.

To construct the datasets, we leverage publicly available datasets to create task-specific datasets. Additionally, we curate an 100k multi-task instruction instruction tuning dataset for complex scenarios using advanced grounding models~\cite{li2024groundinggpt} and GPT-3.5. 

During training, we adopt a three-stage training strategy. Initially, the model is trained to acquire perceptual understanding of multi-modal inputs. Subsequently, it is trained using the task-specific datasets, enabling the model to comprehend human intent, perform reasoning, and effectively accomplish a wide range of tasks. 
Finally, we further fine-tune the model with a multi-turn, multi-task dataset. Inspired by LoRAMOE~\cite{dou2023loramoe}, we incorporate its training methodology to ensure accurate understanding and execution of multiple tasks while mitigating knowledge forgetting and performance degradation. 
Experimental results across multiple multi-modal tasks demonstrate that our model effectively coordinates the MLLM and expert models, outperforming existing methods in task completion.  Furthermore, our unified representation empowers our model to seamlessly integrate more tasks without the need for additional training, further demonstrating the generality and scalability of our approach. In summary, our contributions can be summarized as follows:

\begin{itemize}
    \item We propose a unified representation for multi-task learning by introducing task tokens and grounding tokens to represent different tasks and regions. This enables us to seamlessly integrate multiple tasks.
    \item We construct task-specific datasets and multi-task datasets for complex scenarios. We propose a three-stage training strategy to continuously improve the model's understanding and reasoning abilities while preserving its existing knowledge and capabilities.
    \item Extensive experiments conducted on various benchmarks validate the effectiveness and scalability of our unified approach.The results demonstrate the model's superior performance in handling multiple tasks and its ability to generalize across different domains.
\end{itemize}

\begin{figure*}[t]
    \centering
    \includegraphics[width=\textwidth]{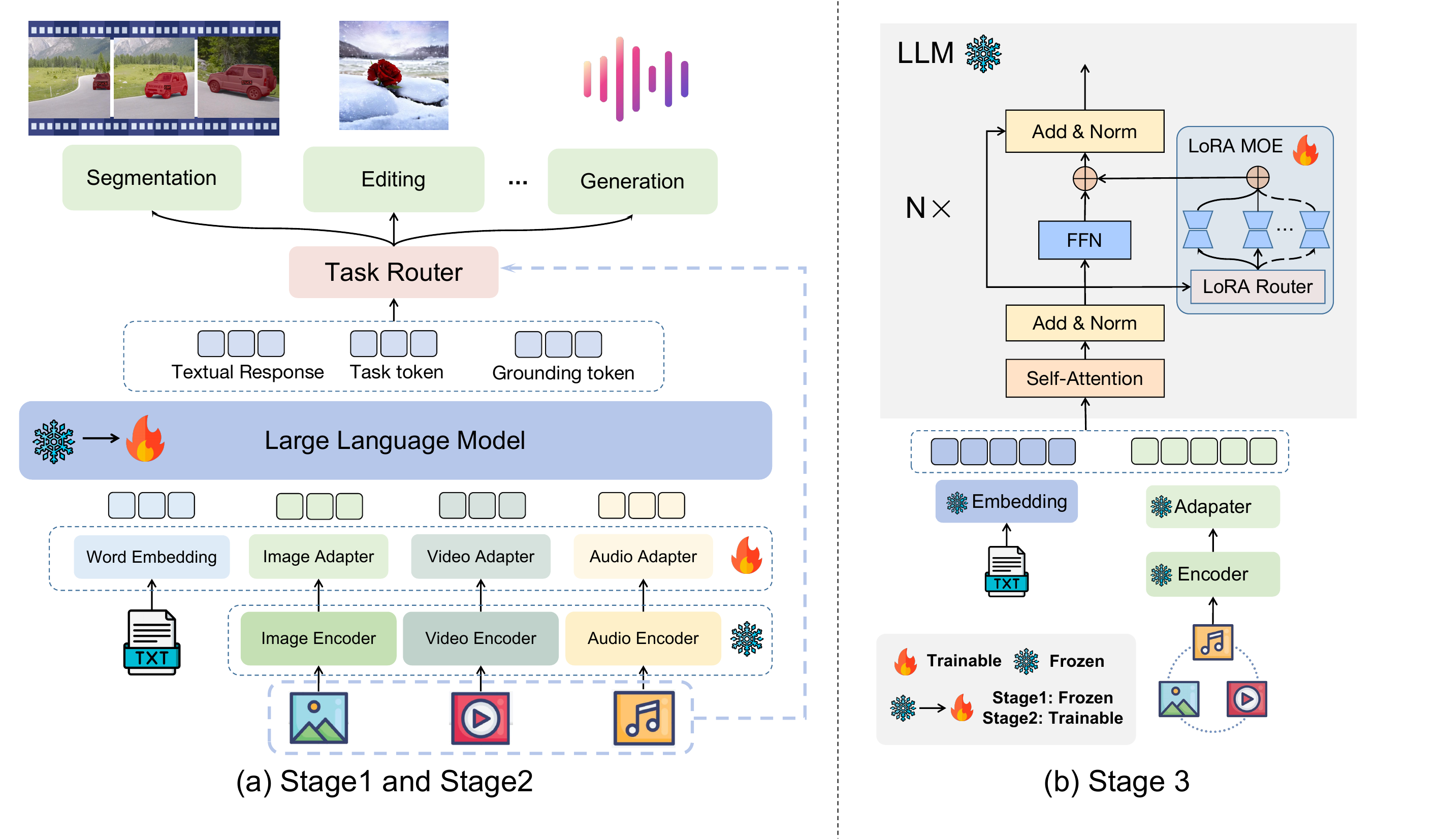}
    \caption{The model structure and three-stage training strategy of UnifiedMLLM.(a) Model structure and training strategy in the first two stages. (b) Training strategy in the third stage.}
    \label{fig:train}
\end{figure*}

\section{Related Work}

\paragraph{Multi-modal Large Language Models (MLLMs)}
In recent years, there has been significant advancements in large language models (LLMs) such as GPTs~\cite{OpenAI2023} and LLaMA~\cite{touvron2023llama} due to their exceptional performance across various natural language processing tasks.
The field of multi-modal language models (MLLMs) has also made notable progress, extending the capabilities of LLMs to handle multi-modal inputs and outputs beyond text alone.

Models like LLaVA~\cite{liu2023visual} and MiniGPT-4~\cite{zhu2023minigpt} have demonstrated remarkable performance in visual question answering tasks. Similarly, video models like Video-LLaMA~\cite{zhang2023video} and Video-Chatgpt~\cite{maaz2023video}, as well as speech models like SpeechGPT~\cite{zhang2023speechgpt}, have also showcased their ability to comprehend the input in multiple modalities.
In MoE-LLaVA~\cite{lin2024moe}, the exploration of incorporating the MOE (Mixture of Experts) structure into MLLMs has yielded outstanding performance while reducing the number of parameters.

\paragraph{Multi-tasks MLLMs}
Some research studies have explored the development of MLLMs capable of handling a greater number of modalities or tasks.
Next-GPT~\cite{wu2023next} achieves multi-modal input and output by connecting modality-specific diffusion models at the output end. 
Trained on multi-modal and multi-granularity data, GroundingGPT~\cite{li2024groundinggpt} is capable of understanding and grounding multi-modal inputs including images, videos, and audios. 
LLaVA-Interactive~\cite{chen2023llava} integrates multiple models and enables tasks such as text-image dialogues, segmentation, generation, and editing, while also facilitating visual interactions. LLaVA-Plus~\cite{liu2023llava} incorporates a skill library comprising various pre-trained visual-language models. It dynamically combines the execution results of these models in real-time based on user's multi-modal inputs to accomplish these tasks. LLMBind~\cite{zhu2024llmbind} integrates different tasks into an MLLM by designing specific tokens. It can handle multi-modal inputs and invoke corresponding models to accomplish various tasks. However, these methods lack uniformity in handling multiple tasks and also do not possess strong capabilities in understanding human intent and reasoning.

\begin{figure*}[t]
    \centering
    \includegraphics[width=\textwidth]{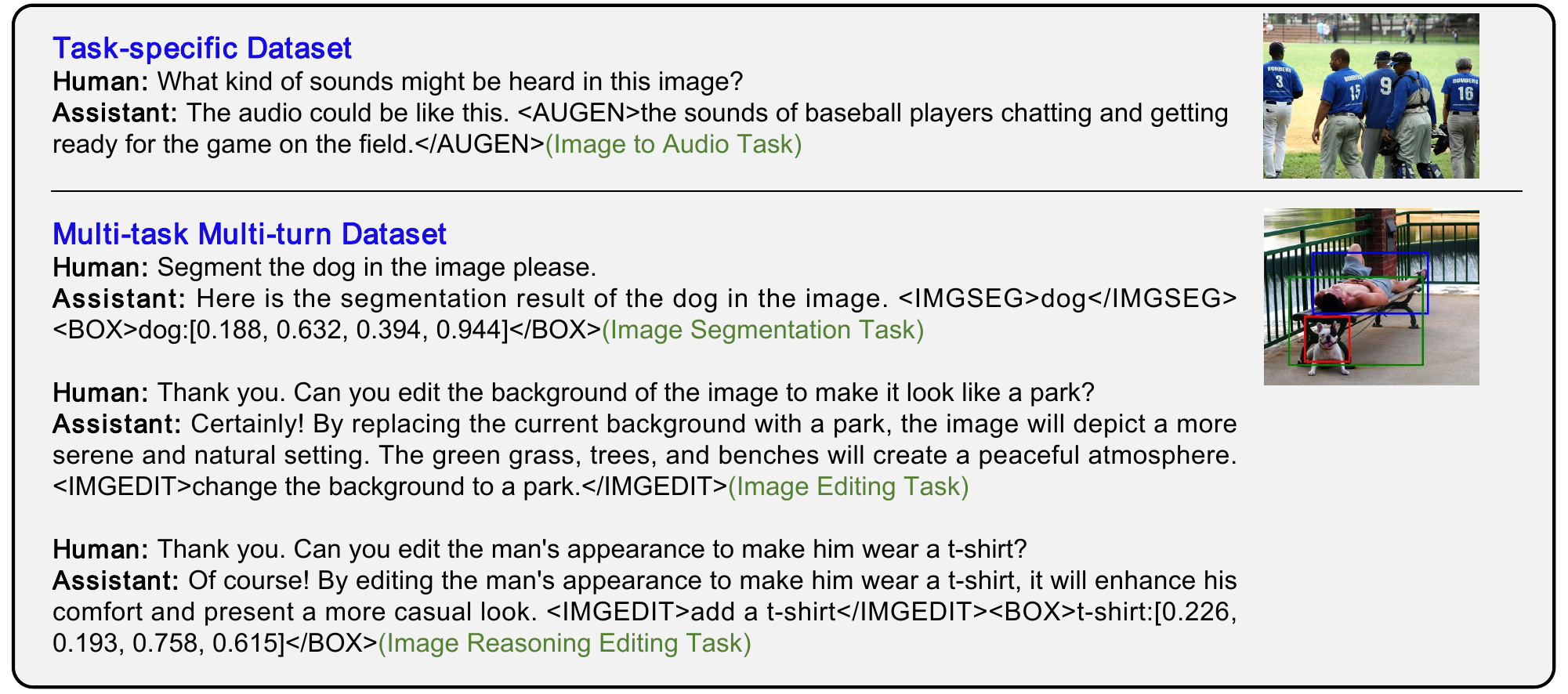}
    \caption{Examples of UnifiedMLLM Dataset in Unified Representation Format. We provide examples of task-specific datasets and multi-task multi-turn datasets.}
    \label{fig:dataset}
\end{figure*}

\section{Method}
We propose UnifiedMLLM, a unified multi-modal model capable of handling various tasks in a unified manner. We will introduce the structure of UnifiedMLLM and its multi-task unified representation. Then we describe the pipeline for constructing the training dataset and our training strategy.

\subsection{Architecture}
Figure~\ref{fig:train} presents the overall architecture of the model. Then we will proceed to introduce each component of the model structure.

\paragraph{Encoder and Adapter}
For each modality input, we employ different encoders to extract features, followed by modality-specific adapters. Specifically, we employ the CLIP visual encoder ViT-L/14~\cite{radford2021learning} to extract image features. For videos, we extract image features by uniformly sampling M frames. After adding temporal positional encoding to the video frames, we aggregate the video features using Q-Former, which has a structure similar to BLIP-2~\cite{li2023blip}. For audio modality, we sample N 2-second audio segments and extract features using the audio encoder from Imagebind~\cite{girdhar2023imagebind}.
Similar to the video branch, a Q-Former is used to aggregate the audio features with the added temporal positional encoding. Following the feature extraction process, each modality input obtains a fixed-length embedding. Then, we use modality-specific adapters, which are two-layer MLPs, to map the features to the embedding space of LLMs.

\paragraph{Unified Representation}\label{Representation}

Due to the different input-output formats across different tasks, achieving a unified approach to scaling and modeling various tasks poses a significant challenge. 

Moreover, for fine-grained tasks like image reasoning editing, it is essential for the model to accurately identify the specific regions to be edited based on the given instruction, as this greatly influences the successful completion of the user's requested task.
As shown in Figure~\ref{fig:train}, in addition to generating textual responses, our language model generates task tokens and grounding tokens. To achieve this, we expand the vocabularies of the LLM and introduce multiple task-specific tokens and grounding tokens, which appear in pairs ({e.g., <Edit></Edit>). The content between task tokens indicates the task to be executed, while the content between grounding tokens contains region-relative coordinates expressed in text format.
To handle the various tasks and modalities, we employ a task router component, which utilizes the special tokens to determine the type and region of the task to be executed. The task router will activate the corresponding expert model to perform the task based on the special tokens.
This representation approach facilitates seamless integration of different tasks across multiple modalities. Furthermore, decoupling the LLM from the subsequent expert models not only reduces training costs but also ensures excellent scalability.

\paragraph{Experts Integration}

We activate different external modules based on the output of the task router to execute different tasks, enabling seamless integration of various tasks.  
For text-to-image generation and layout-based generation tasks, we utilize Stable Diffusion~\cite{rombach2022high} and GLIGEN~\cite{li2023gligen} models. 
Instructpix2pix~\cite{brooks2023instructpix2pix} and GLIGEN are employed for image global editing and reasoning editing tasks in image editing. 
SEEM~\cite{zou2024segment} is utilized for image and video segmentation tasks. 
For video editing tasks, we utilize the FRESCO~\cite{yang2024fresco} model. ModelScopeT2V~\cite{wang2023modelscope} and I2vgen-xl~\cite{zhang2023i2vgen} are used for text-based video generation and image-based video generation, respectively. Additionally, Auffusion~\cite{xue2024auffusion} is employed for audio generation.

\paragraph{LoRA Mixture of Experts}

It has been observed~\cite{dou2023loramoe} that when LLMs introduce a large amount of instruction data during the SFT stage to enhance performance on multiple tasks, it may compromise the stored world knowledge within the model. In order to mitigate this issue and ensure that the model retains its reservoir of knowledge and reasoning abilities during the training process, we adopt a strategy where the backbone of the model is frozen to preserve its capabilities. Additionally, multiple expert models are introduced to handle various downstream tasks. We employ LoRA as the structure for the expert models, which enhances the efficiency of both training and inference processes.
For the transformers architecture, the forward propagation of the feed-forward neural (FFN) network block can be denoted as follows:
\begin{equation}
    f(x)=x+f_{\mathrm{FNN}}(x).
\end{equation}
The linear layer in the FFN can be expressed as:
\begin{equation}
    o=Wx=W_{0}x+\Delta Wx
\end{equation}
where $W_{0}\in\mathbb{R}^{d_{\mathrm{in}}\times d_{\mathrm{out}}}$ represents the parameter of the backbone while $\Delta W\in\mathbb{R}^{d_{\mathrm{in}}\times d_{\mathrm{out}}}$ denotes the updated parameter during training.
We replace the linear layer with the MoE, then the forward process of the layer can be denoted as follows:
\begin{equation}
    \begin{gathered}
        o=W_{0}x+\Delta Wx=W_{0}x+\sum_{i=1}^{N}G(x)_{i}E_{i}(x),\\
        G(x)=Softmax(x\cdot W_{g}),
    \end{gathered}
\end{equation}
where $E_{i}(\cdot)$ denotes the $i$-th expert, $G(\cdot)$ represents the router network and the $W_g$ is a trainable parameter of the route network. With this design, the experts are able to efficiently handle different tasks through collaboration.

To enhance the training efficiency, we replace the experts in the MoE layer with a low-rank format. The parameter matrix 
$\Delta W_E\in\mathbb{R}^{d_{in}\times d_{out}}$ of a single expert can be expressed as follows:
\begin{equation}
    \vartriangle W_{E}=BA,
\end{equation}
where $A\in\mathbb{R}^{d_{in}\times r},B\in r\times\mathbb{R}^{d_{in}}$ and the rank $r<<min(d_{in},d_{out}).$
The forward process of the LoRAMoE layer can be written as follows:
\begin{equation}
    o=W_{0}x+\frac{\alpha}{r}\sum_{i=1}^{N}w_{i}B_{i}A_{i}x,
\end{equation}
where $w_i$ denotes the weight of $i$-th expert and $\alpha$ is a constant.
This low-rank design significantly reduces training costs, improves training speed, and avoids degradation of model knowledge and capabilities during the training process.

\subsection{Dataset}
\paragraph{Task-specific Dataset}

To enable the model to handle different tasks in a unified manner, we construct task-specific datasets following the representation method described in section~\ref{Representation}. For each task, we select task-relevant datasets and transform them into a conversation format, where the model's output includes task tokens and grounding tokens. 
Additionally, to further enhance the model's reasoning ability, we utilize several reasoning datasets constructed in our work. These include the reasoning segmentation dataset from LISA~\cite{lai2023lisa}, the reasoning editing dataset from SmartEdit~\cite{huang2023smartedit}, and the layout-based image generation dataset from LayoutGPT~\cite{feng2024layoutgpt}. These datasets further enhance the model's understanding of human intent. Figure~\ref{fig:dataset} showcases some task-specific datasets.

\paragraph{Multi-turn Multi-task Dataset}
The existing multi-task datasets are quite limited, especially those with coordinates for regions like the reasoning segmentation dataset. 
Due to our model's unified representation for different tasks, we can maximize the model's performance across a wider range of tasks, even with limited data availability.
To further expand our dataset with grounding tokens, we utilize advanced grounding model GroundingGPT~\cite{li2024groundinggpt} for data generation. Given an input image, we first use GroundingGPT to generate captions with bounding boxes. Then we utilize GPT-3.5 for multi-turn dialogue data construction. By providing GPT-3.5 a system prompt that outlines roles, requirements, and several human-annotated examples, we ask GPT-3.5 to generate multi-turn dialogues using the provided captions. Subsequently, we filter the generated data to remove samples that do not adhere to the expected output format. We totally generate 100k instances of multi-turn, multi-task dialogues, covering various multi-modal tasks in complex scenarios.

\begin{table*}[t]
    \centering
    \setlength\tabcolsep{2mm}
    \begin{tabular}{c|ccc|ccc|cc}
    \toprule
    \multirow{2}{*}{ Method } & \multicolumn{3}{|c|}{ RefCOCO } & \multicolumn{3}{c|}{ RefCOCO+ } & \multicolumn{2}{c}{ RefCOCOg } \\
    \cline { 2 - 9 } & val & testA & testB & val & testA & testB & val & test\\
    \midrule

    VLT~\cite{ding2021vision}  & 67.5 & 70.5 & 65.2 & 56.3 & 61.0 & 50.1 & 55.0 & 57.7 \\
    CRIS~\cite{wang2022cris}  & 70.5 & 73.2 & 66.1 & 62.3 & 68.1 & 53.7 & 59.9 & 60.4 \\
    LAVT~\cite{yang2022lavt}  & 72.7 & 75.8 & 68.8 & 62.1 & 68.4 & 55.1 & 61.2 & 62.1 \\
    LISA~\cite{lai2023lisa}  & 74.9  &  \textbf{79.1}  & 72.3  & 65.1 & 70.8 & 58.1  & 67.9 & \textbf{70.6}  \\
    NExT-Chat~\cite{zhang2023next} & 74.7  &  78.9  & 69.5  & 65.1 & 71.9 & 56.7  & 67.0 & 67.0  \\
    \midrule
    UnifiedMLLM  & \textbf{76.3} & 78.8 & \textbf{72.7} & \textbf{66.4} & \textbf{72.4} & \textbf{59.1} & \textbf{68.0} & 69.6 \\
    \bottomrule
    \end{tabular}
    \caption{Quantitative results of image referring image segmentation on three referring segmentation datasets: RefCOCO, RefCOCO+, and RefCOCOg with metric cIoU.}
    \label{tab:seg}
\end{table*}

\begin{table*}[t]
    \centering
    \setlength{\tabcolsep}{1mm}{
    \begin{tabular}{c|c|c|c|c|c|c}
    \toprule
    \multirow{2}{*}{Methods} & \multicolumn{3}{c|}{Understanding Scenarios} & \multicolumn{3}{c}{Reasoning Scenarios} \\
    \cline{2-7} & PSNR$\uparrow$ & SSIM$\uparrow$  & CLIP Score$\uparrow$  & PSNR$\uparrow$ & SSIM $\uparrow$ & CLIP Score$\uparrow$ \\
    \midrule
    InstructPix2Pix~\cite{brooks2023instructpix2pix} & 21.576 & 0.721 & 22.762  & 24.234 & 0.707 & 19.413 \\
    MagicBrush~\cite{zhang2024magicbrush} & 18.120 & 0.68 & 22.620 & 22.101 & 0.694  & 19.755 \\
    InstructDiffusion~\cite{geng2024instructdiffusion} & \textbf{23.258} & 0.743  & 23.080  & 21.453 & 0.666 & 19.523  \\
    SmartEdit~\cite{huang2023smartedit} & 22.049 & 0.731  & 23.611 & 25.258 & 0.742 &  20.950\\
    \midrule
    UnifiedMLLM & 20.670 & \textbf{0.776}  & \textbf{23.633} & \textbf{26.667} & \textbf{0.808}  & \textbf{21.104} \\
    \bottomrule
    \end{tabular}}
    \caption{Quantitative results (PSNR/SSIM/CLIP Score) of reasoning editing on Reason-Edit~\cite{huang2023smartedit}.}
    \label{tab:edit}
\end{table*}

\subsection{Training}
We adopted a three-stage training strategy. Firstly, we train the model to acquire the ability to perceive and understand different modal inputs. Secondly, we train the model using multiple task-specific datasets to develop its capability to understand human intent and complete different tasks. Lastly, we further optimize the model's responses and enhance its reasoning ability to enable it to complete a variety of tasks in complex scenarios.
\paragraph{Modality-perception Pretraining}
In this stage, we expect the model to understand multi-modal inputs and establish the knowledge base, which serves as the foundation for subsequent reasoning and completion of various multi-modal tasks. During training, we utilize publicly available multi-modal training data, consisting of three pre-training datasets for each modality.
Throughout the training process, we keep the LLM and encoder frozen and only train the adapters for each modality.

\paragraph{Task Adaptation Tuning}
After the the first stage of training, where the model gains the ability to understand inputs, it still lacks the capability to handle various multi-modal tasks. In this stage, we train the model to understand human intent and accomplish a variety of tasks. The training data used in this stage includes task-specific datasets that we constructed based on publicly available data, following the unified representation format described in section~\ref{Representation}. These datasets contain replies with task tokens and grounding tokens, enabling the model to comprehend human intent.
Additionally, we also use some open source general instruction fine-tuning datasets for training to improve the model's ability to understand general instructions.
During this stage of training, we keep the encoders for each modality frozen while jointly training the LLM and adapters.

\paragraph{Multi-task LoRAMoE Tuning}
To enable the model to further understand human intent, perform reasoning, and accomplish a variety of tasks in complex scenarios while avoiding knowledge forgetting and performance degradation caused by further training, we utilize the constructed multi-turn multi-task dataset for training. As depicted in Figure~\ref{fig:train}, during the training process, we keep all parameters frozen except for LoRAMOE, which is updated. This training strategy enhances the model's capability to handle different tasks in complex scenarios while preserving its general ability and maintaining training efficiency.

\section{Experiment}
\begin{table*}[t]
    \centering
    \setlength\tabcolsep{0.5mm}
    \begin{tabular}{c|ccccc|ccc}
    \toprule 
    \multirow[c]{3}{*}{ Methods } & \multicolumn{5}{|c|}{ Numerical Reasoning } & \multicolumn{3}{|c}{ Spatial Reasoning } \\
    \cline{2-9} & \multicolumn{3}{c}{ Layout Eval. } & \multicolumn{2}{c|}{ Image Eval. } & Layout Eval & \multicolumn{2}{c}{ Image Eval.} \\
    \cline{2-9} & Precision & Recall & Acc & Acc-G & Sim-C & Acc & Acc-G & Sim-C \\
    \midrule

    Stable Diffusion (v2.1) & - & - & - & 42.44 & 0.256 & - & 17.81 & 0.256 \\

    Attend-and-Excite~\cite{chefer2023attend} & - & - & - & 45.74 & 0.254 & - & 26.86 & 0.264 \\

    LayoutTransformer~\cite{gupta2021layouttransformer} & 75.70 & 61.69 & 22.26 & 40.55 & 0.247 & 6.36 & 28.13 &  0.241 \quad  \\
    LayoutGPT (GPT-3.5)~\cite{feng2024layoutgpt}& \textbf{94.81} & \textbf{96.49} & \textbf{86.33} & 51.20 & 0.258 & 82.54 & 52.86 & 0.264 \\

    LayoutGPT (GPT-4) & 78.36 & 86.29 & 78.43 & 55.64 & 0.261 & 91.73 & 60.64 & 0.268 \\
    \midrule
    UnifiedMLLM & 93.03 & 95.02 & 85.43 & \textbf{57.94} & \textbf{0.266} & \textbf{92.93} & \textbf{61.78} & \textbf{0.270}\\

    \bottomrule
    \end{tabular}
    \caption{Quantitative results of layout-guided image generation on NSR-1K~\cite{feng2024layoutgpt}, evaluated using counting and spatial correctness. "Acc-G" refers to accuracy calculated based on the GLIP~\cite{li2022grounded} model, while "Sim-C" refers to similarity calculated based on the CLIP~\cite{radford2021learning} model. }
    \label{tab:layout}
\end{table*}

\begin{table}[t]
    \centering
    \setlength\tabcolsep{5mm}
    \begin{tabular}{cc}
    \toprule
    Method & FID $\downarrow$\\
    \midrule
    GLIDE~\cite{nichol2021glide} & 12.24\\
    GILL~\cite{koh2024generating} & 12.20\\
    Emu~\cite{sun2023generative} & 11.66\\
    Codi~\cite{tang2024any} & 11.26\\
    NExT-GPT~\cite{wu2023next} & 11.28\\
    \midrule
    UnifiedMLLM  & \textbf{10.84} \\
    \bottomrule
    \end{tabular}
    \caption{Quantitative results of text-to-image generation on COCO-captions
dataset, evaluated with FID.}
    \label{tab:t2i}
\end{table}

\begin{table}[t]
    \centering
    \setlength\tabcolsep{0.0mm}
    \begin{tabular}{ccc}
    \toprule
    Method & FID$\downarrow$ & CLIPSIM$\uparrow$\\
    \midrule
    CogVideo~\cite{hong2022cogvideo} & 23.59 & 0.2631\\
    Make-A-Video~\cite{singer2022make} & 13.17 & 0.3049\\
    Latent-Shift~\cite{an2023latent} & 15.23 & 0.2773\\
    NExT-GPT~\cite{wu2023next} & 13.04 & 0.3085\\
    \midrule
    UnifiedMLLM  & \textbf{11.15} & \textbf{0.3120} \\
    \bottomrule
    \end{tabular}
    \caption{Quantitative results of text-to-video generation on MSR-VTT
dataset, evaluated with FID and CLIPSIM.}
    \label{tab:t2v}
\end{table}

\begin{table}[t]
    \centering
    \begin{tabular}{ccc}
    \toprule
    Method & FD$\downarrow$ & IS$\uparrow$\\
    \midrule
    DiffSound~\cite{yang2023diffsound} & 47.68 & 4.01\\
    AudioLDM-S~\cite{liu2023audioldm} & 29.48 & 6.90\\
    AudioLDM-L~\cite{liu2023audioldm} & 23.31 & 8.13\\
    NExT-GPT~\cite{wu2023next} & 23.58 & 8.35\\
    \midrule
    UnifiedMLLM  & \textbf{22.42} & \textbf{9.95} \\
    \bottomrule
    \end{tabular}
    \caption{Quantitative results of text-to-audio generation on AudioCaps dataset, evaluated with FID and IS.}
    \label{tab:t2a}
\end{table}

\subsection{Experimental Setup}
We employ Vicuna-v1.5~\cite{chiang2023vicuna} as the language model. Each training stage lasts for one epoch. During the training process, all images were padded to a square shape and resized to a resolution of 336 × 336. For each video, 64 frames were sampled, and for each audio, three 2-second segments were sampled and processed. All experiments were conducted on 8 A100-80G GPUs.

\subsection{Quantitative Evaluation}
\paragraph{Referring Segmentation}
For the reference segmentation task, the model needs to segment the objects in the image corresponding to the given expressions. We conduct experiments using the RefCOCO~\cite{kazemzadeh2014referitgame}, RefCOCO+~\cite{kazemzadeh2014referitgame}, and RefCOCOg~\cite{mao2016generation} datasets and evaluate the models based on the cIoU metric. 

As shown in Table 2, we have achieved excellent results on multiple datasets due to the strong grounding ability of our model.

\paragraph{Reasoning Editing}
The image reasoning edit task requires the model to reason the areas that need editing based on user instructions and perform editing. We conducted experiments on the Reason-Edit~\cite{huang2023smartedit} dataset. For the background, we evaluated the models using the PSNR and SSIM metrics. For the foreground regions that require editing, we calculated the CLIP Score between the edited foreground regions in the image and the ground truth labels. The results are shown in Table~\ref{tab:edit}.
It can be observed that our method achieves better results in both understanding and reasoning the scenes compared to other methods. Our model successfully edits the target regions while avoiding interference with the background areas.

\paragraph{Layout-based Image Generation}
The layout-based image generation task is used to evaluate the controllable generation capability of the model,where the objective was to generate coherent images by arranging the layout based on user instructions. Evaluations are conducted using the NSR-1K~\cite{feng2024layoutgpt} dataset to examine the model's proficiency in comprehending quantity and spatial relationships for layout tasks. Following LayoutGPT~\cite{feng2024layoutgpt}, for the numerical reasoning subset, we report precision, recall, and accuracy based on generated bounding box counts and spatial positions. For spatial seasoning, we use the bounding box center for evaluation. For evaluating the generated images, we use GLIP~\cite{li2022grounded} to obtain bounding boxes and compute average accuracy based on the bounding box counts or spatial relations. Additionally, we also report the CLIP cosine similarity between text prompts and generated images. 
As shown in Table~\ref{tab:layout}, our model is capable of generating layouts that are more reasonable and accurate compared to other models. Additionally, the generated images exhibit better consistency with the prompts, validating the reasoning and planning ability of our model.
\paragraph{Multi-modality Generation}
 In this section, we evaluate the performance of our model in multi-modal text-based generation tasks. Specifically, in the text-to-image generation task, we evaluate using the COCO-caption~\cite{lin2014microsoft} dataset, and the evaluation metric is the Fréchet Inception Distance (FID) score. In the text-to-video generation task, we evaluate using the MSR-VTT~\cite{xu2016msr} dataset, and the evaluation metrics are FID for content quality  and CLIPSIM for textual alignment. Furthermore, in the text-to-audio generation task, we conduct experiments on the AudioCaps~\cite{kim2019audiocaps} dataset and evaluate using the Frechet Distance (FD) and Inception Score (IS) metrics. As observed from Tables~\ref{tab:t2i},~\ref{tab:t2v},~\ref{tab:t2a}, compared to previous expert models or multi-task models, our model demonstrates strong performance across various multi-modal generation tasks.

\subsection{Qualitative Results}

Figure~\ref{fig:demo} presents a selection of visual results that effectively demonstrate the remarkable capabilities of our model. These results demonstrate our model's exceptional performance in tasks involving multi-modal understanding, segmentation, generation, editing and so on.
As depicted in the Image Editing example, our model is able to comprehend and reason implicit human intent, enabling it to select the appropriate regions for editing. Additionally, our model exhibits robust generalization capabilities, successfully completes tasks that it has not encountered during training. For instance, tasks such as generating videos from images and audio, as depicted in the bottom right of the figure, validate the scalability of our model.


\section{Conclusion}
In this paper, we propose UnifiedMLLM, a multi-modal large language model that handles various multi-modal tasks using a unified representation. By introducing task tokens, grounding tokens, and a task router, we seamlessly integrate multiple tasks with excellent scalability and versatility. We construct a task-specific dataset and a multi-task multi-turn instruction-tuning dataset, and employ a three-stage training approach to enable the model to effectively perform diverse multi-modal tasks while avoiding degradation of general capabilities. Due to the powerful reasoning and grounding abilities of our model, a significant number of quantitative experiments and visual results demonstrate the effectiveness of our approach.

\section{Limitation}
\paragraph{Model Architecture}
Due to limited training resources and the complexity of tasks, our model primarily relies on external models to accomplish various multi-modal tasks. This approach ensures the effectiveness and scalability of completing visual tasks. However, the scope and effectiveness of the model are still constrained by the expert models. A future research direction is to construct an end-to-end trainable multi-modal system. One possible approach is to discretize various modal information, following the methodology of AnyGPT~\cite{zhan2024anygpt}.

\paragraph{Multi-modal Interleaving}
Currently, our model mainly focuses on processing single-modal inputs. Effectively handling multi-modal information simultaneously or interleaved is a challenge that needs to be addressed. CoDi-2~\cite{tang2024codi} provides some insights, but due to the lack of this type of data, the number of tasks that can be handled is relatively limited. A future research direction is to explore how to achieve interleaved understanding and generation of inputs and outputs.

\bibliography{anthology,custom}
\bibliographystyle{acl_natbib}

\appendix

\end{document}